\documentclass[runningheads]{llncs}
\usepackage{csquotes}
\usepackage{graphicx}
\begin{document}

\title{Augmenting Holistic Review in University Admission using Natural Language Processing for Essays and Recommendation Letters
}
%
%
\author{Jinsook Lee\inst{1}\orcidID{0000-0002-9957-1342} \and
Bradon Thymes\inst{1}\orcidID{0000-0002-3754-404X} \and 
Joyce Zhou\inst{1}\orcidID{0000-0003-1205-3970} \and
Thorsten Joachims\inst{1}\orcidID{0000-0003-3654-3683} \and
René F. Kizilcec\inst{1}\orcidID{0000-0001-6283-5546}}
\authorrunning{J. Lee et al.}
%
\institute{Cornell University, Ithaca NY 14850, USA}
\maketitle              
\begin{abstract}
University admission at many highly selective institutions uses a holistic review process, where all aspects of the application, including protected attributes (e.g., race, gender), grades, essays, and recommendation letters are considered, to compose an excellent and diverse class. In this study, we empirically evaluate how influential protected attributes are for predicting admission decisions using a machine learning (ML) model, and in how far textual information (e.g., personal essay, teacher recommendation) may substitute for the loss of protected attributes in the model. Using data from 14,915 applicants to an undergraduate admission office at a selective U.S. institution in the 2022-2023 cycle, we find that the exclusion of protected attributes from the ML model leads to substantially reduced admission-prediction performance. The inclusion of textual information via both a TF-IDF representation and a Latent Dirichlet allocation (LDA) model partially restores model performance, but does not appear to provide a full substitute for admitting a similarly diverse class. In particular, while the text helps with gender diversity, the proportion of URM applicants is severely impacted by the exclusion of protected attributes, and the inclusion of new attributes generated from the textual information does not recover this performance loss.

\end{abstract}
\section{Introduction}
As applications of machine learning (ML) and artificial intelligence (AI) in education have become more widespread, there has been increased attention to questions of algorithmic bias and fairness~\cite{kizilcec2020algorithmic,baker2021algorithmic,holmes2022ethics,woolf2022introduction}. In particular, the question of whether protected attributes should be considered in such models and the consequences of their omission has been debated in the literature~\cite{bakerusing,yu2021should,yu2020towards}. This question directly addresses an important policy issue in the context of holistic review in undergraduate admission. At many selective institutions of higher education, admission offices have adopted a process known as holistic review, where applications are read in their entirety to contextualize an applicant's academic record, extracurricular activities, personal essays, and teacher recommendations~\cite{stevens2007creating}. The holistic review process has historically included applicants' protected attributes, such as their gender and race, to take their social identity into consideration and compose a diverse class. However, consideration of protected attributes in holistic admission may be halted following legal challenges to the practice of affirmative action~\cite{santoro2022us}.

Holistic review in undergraduate admission at selective institutions is a time-constrained and cognitively complex process that does not scale well with increasing numbers of applicants and fewer standardized measures due to a recent move towards test-optional admission policies~\cite{bennett2022untested}. This has motivated research over the past decade to examine potential applications of ML in admission to support experienced admission professionals in accomplishing the task of selecting applicants to admit~\cite{lee2023evaluating,ragab2012hrspca,sridhar2020university,vaghela2015students}. Recent work has examined how different components of an application (e.g., protected attributes, standardized test scores) influence model performance and the demographic distribution of predictions~\cite{lee2023evaluating}. Another line of research has examined the socio-linguistic properties of textual data in the application, such as personal essays, using Natural Language Processing (NLP) and Large Language Models (LLMs)~\cite{alvero2020ai,alvero2021essay,lirausing,pennebaker2014small}. Together, this research indicates that omitting protected attributes has negative consequences for model performance and diversity, and textual information strongly correlates with applicants' socio-demographic characteristics. However, prior research has not examined these factors simultaneously to understand how the omission of protected attributes and the addition of textual information influences model performance and prediction diversity.

In this study, we use data from 14,915 applicants to an undergraduate admission office at a selective U.S. institution in the 2022-2023 application cycle to evaluate the influence of omitting protected attributes (gender, race/ethnicity) and of including different kinds of textual information (student essay, writing supplements, teacher recommendation letters) as inputs for ML models that predict admission decisions. We evaluate the different feature sets in terms of the overall model performance as well as the characteristics of the applicants the model ranks in the top decile for admission (\enquote{top pool}). This study provides novel insight into the consequences of omitting protected attributes in a learned admission prediction model for performance and diversity, and to what extent the inclusion of textual application materials in the model can recover the expected loss in the diversity of the pool of highly ranked applicants.

\section{Related Work}

Many university admission essay prompts invite applicants to share insights into their background, experiences, and identity, as these aspects play a significant role in the holistic review process~\cite{stevens2007creating}. Recognizing the richness of information contained within these essays, researchers have turned to NLP techniques to extract socio-linguistic information and analyze resulting patterns. Prior work analyzed the content and writing styles of college admission essays by applying Correlational Topic Modeling (CTM) and Linguistic Inquiry and Word Count (LIWC) and found that the essays are closely correlated to SAT scores and household income~\cite{alvero2021essay}. Studies that use embedding representations have shown that word vectors trained on essays from students with lower household incomes exhibit comparatively poorer performance on tasks assessing semantic similarity and analogy when compared to those derived from essays written by students from higher-income backgrounds. These results suggest that socioeconomic factors can manifest in the content and style of college admission essays~\cite{arthurs2020whose}. Other applications of NLP have tested prediction models trained on manually coded features of essays to scale up labels that describe extracurricular and work experiences~\cite{lirausing}.

ML approaches have gained prominence in college admission research as a means of augmenting the complex work that admission professionals perform. For example, Lee and colleagues~\cite{lee2023evaluating} evaluated the potential of using a trained model instead of standardized test scores to organize a large pool of applicants into ten smaller pools based on the predicted probability of admission. The findings revealed that the ML approach surpassed an SAT-based heuristic in terms of performance and it more accurately aligned with the demographic makeup of the previously admitted applicant pool. In a similar context, another study evaluated a learned model to predict the acceptance likelihood of admitted applicants to help universities estimate the expected size of the incoming student cohort~\cite{basu2019predictive}.


Despite an increasing number of efforts to evaluate ML and NLP techniques in university admission processes, most prior research considers a limited scope of application and decision data as inputs in the modeling process. Many studies that consider linguistic features of essays do not have final admission decisions as an outcome to evaluate model accuracy or detailed socio-demographic attributes of applicants to examine model bias. This study addresses this gap in the literature by examining textual application materials in conjunction with demographic features and admission outcomes in a large dataset. This work examines the specific contributions of essays and recommendation letters within the broader context of the admission process. We analyze the relative importance of these texts and assess their potential to enhance the performance of a learned admission prediction model. Our goal is to understand how text analysis can be used to help admission professionals in composing a diverse undergraduate class.
\raggedbottom

\section{Methods}

\subsection{Context}
The case university is a selective higher educational institution in the United States. The admission process at the case university has two major cycles, the Early Decision (ED) and the Regular Decision (RD) cycle. In the 2022-2023 admission period, there were 1,954 ED applicants with a 22.8\% acceptance rate and 16,139 RD applicants with a 4.9\% acceptance rate. The ED admission rate is typically higher because applicants are not supposed to enroll in other institutions if admitted during the ED cycle.

All first-year applicants are required to apply to the case institution via the Common Application portal. Applications contain various information about applicants, including  SAT or ACT scores, AP and/or IB scores, TOEFL/IELTS scores, honors and awards, descriptions of extracurricular activities, and one personal essay (in response to one of seven U.S.-wide prompts). Additionally, at the case university, applicants are required to submit two recommendation letters and two writing supplements, which are college-specific essays. One writing supplement has a fixed prompt, while the other provides applicants with a choice of two prompts, one of which invites the applicant to discuss their identity. The writing supplements and the teacher recommendation letters were saved in PDF format.

\subsection{Dataset and Pre-Processing}
We obtained a de-identified set of 18,093 applications received in the 2022-23 admission period for matriculation in Fall 2023. We omitted incomplete applications and dummy data. This results in 17,417 applications available in the analysis. We created 303 features based on the data extracted from the Common Application form (excluding essays, writing supplements, and recommendations). We extracted the text data from the PDF files of the writing supplement and teacher recommendation letters. For 14.4\% of the sample, either the writing supplement or the teacher evaluations were missing from the application file or could not be extracted from the PDF. To ensure consistency in training data for model comparisons, we excluded applications without these texts. All models were trained on the same 14,915 applications.

All text was pre-processed with the \textit{NLTK} library~\cite{bird2009natural} and we applied Porter stemming and removed non-alphabet characters. We used two conceptually different approaches to represent the textual materials quantitatively. First, we convert the text data to word vectors using a bag-of-words approach with TF-IDF weighted unigrams. This is a common approach to represent low-level socio-linguistic patterns~\cite{pennebaker2014small}. Second, we computed topic models via Latent Dirichlet Allocation $($LDA$)$ \cite{blei2003latent} using the \textit{Gensim} library \cite{rehurek2011gensim}. It is commonly utilized to represent higher-level socio-linguistic patterns and themes, also used in prior work on admission essays~\cite{alvero2021essay}. We created 20 topics for each document: the essay, the writing samples, and the recommendation letters. We explored increasing the number of topics but found no notable improvements. We combine the resulting vectors from both the TF-IDF and LDA approach when training models with text.

We used a bag-of-words approach to represent the text instead of more sophisticated language models as an initial step because prior research with college admission essays found minimal performance gains from using Deep Neural Network (DNN) compared to Multinomial Naive Bayes, which also uses word frequencies to predict applicants' demographic attributes~\cite{alvero2020ai}.

\subsection{Modeling}

We use probabilistic binary classification to predict college admission: admitted applicants have a label of \enquote{admitted} or \enquote{conditionally admitted} in the data, whereas denied applicants have a label of \enquote{waitlisted}, \enquote{withdrawn}, or \enquote{rejected}. We randomly sampled 80\% of the dataset for training (n=11,932) and the remaining 20\% was used for testing the model (n=2,983). For training, we fit a Gradient Boosting Decision Trees model using the \textit{Sklearn} library with its default parameter setting.

In this study, we trained ten different models by varying the combinations of textual features (no text, all text, personal essay only, writing supplements only, teacher recommendation letters only) and demographic features (gender/race vs. none of them). Our analysis included 13 features relevant to gender and race based on questions in the Common App application form, namely Sex, Gender Identity, URM, Hispanic Latino background, African background, American Indian background, Asian background, White background, and Race (American Indian or Alaska Native, Asian, White, Black or African American, Native Hawaiian or Other Pacific Islander). The 10 models were based on different combinations of datasets, with a focus on two key factors: the inclusion of the 13 demographic features, and the inclusion of text data (personal essays, writing supplements, and teacher recommendation letters). Our baseline model was trained on the Common App dataset with demographic features but without textual features.

For evaluation purposes, we calculated for each model (1) AUC ROC scores as a measure of overall model performance, (2) the proportion of admitted applicants in each applicant pool as a more specific measure of performance, and (3) the proportions of female and of URM applicants predicted to be in each applicant pool as a measure of diversity. Applicant pools are defined based on deciles of the predicted probability of admission, such that pool 1 comprises 10\% of applicants with the lowest predicted admission rate, while pool 10 comprises the 10\% of applicants with the highest predicted admission rate (adopting the approach from~\cite{lee2023evaluating}).

\section{Results}

\subsection{Overall Model Performance}
Table \ref{tab1} shows the performance in terms of AUC for each model when demographic information is added/removed (columns) and with different sets of textual materials included (rows). As expected, we find that removing demographic information from the baseline model without text results in a large drop in model performance (permutation test: $p<0.001$). Adding all textual features provides surprisingly minimal gains in prediction performance when demographic information is present ($p=0.303$). The teacher recommendation letters alone provide no improvement in performance at all.

Among the models without demographic information, adding all of the text features improves the model performance, though not significantly ($p=0.102$), and it still does not reach the level of performance of any of the models with demographic features ($p<0.001$). Of all the texts, the writing supplement appears to be most effective at partially recovering the drop in model performance due to omitting demographic information ($p=0.045$). This may be a reflection of demographic information encoded in the applicants' responses to the identity-related prompts of the writing supplement. In contrast, model performance is slightly reduced with the addition of just the teacher recommendation letters ($p=0.046$), or just the personal essay, though not significantly ($p=0.253$).

\begin{table}[!]
\centering
\caption{Performance predicting admission decisions (AUC ROC) with models trained on different combinations of textual and demographic features.}\label{tab1}
\begin{tabular}{|l|c|c|}
\hline
 & \multicolumn{2}{c|}{\textbf{ Demographics }} \\
\textbf{Textual Features} &  Excluded  &  Included \\ \hline
No text             	&       0.809 	    &      0.884    \\ \hline
All text            	&       0.821       &      0.892    \\ \hline
Personal essay      	&       0.804     	&      0.887   	\\ \hline
Writing supplements 	&       0.829     	&      0.887  	\\ \hline
Teacher recommendation letter & 0.796     	&      0.884   	\\ \hline
\end{tabular}
\end{table}

\subsection{Model Performance and Diversity by Applicant Pool}

We next examined the demographic distribution in terms of gender and underrepresented minority (URM) status within applicant pools. Table~\ref{tab2} presents the percentages of female applicants, URM applicants, and admitted applicants in the top 10\% pool (i.e., applicants with the highest predicted probability of admission). In models without textual features, the omission of demographic features causes a significant drop in the top pool in prediction performance (i.e., percentage of admitted applicants: $\chi^2(df=1)=7.69, p=0.005$) and diversity in terms of the percentage of URM applicants $(\chi^2(df=1)=142.32, p<0.001)$ but not the percentage of female applicants $(\chi^2(df=1)=0.18, p=0.67)$.

The addition of textual features to a model with demographic features does not significantly improve model performance ($\chi^2(df=3)=0.21, p=0.98$) or diversity in the top pool (URM: $\chi^2(df=3)=0.85, p=0.83$; female: $\chi^2(df=3)=0.30, p=0.96$). However, for models without demographic features, adding textual information substantially recovers the percentage of women in the top pool, though it does not recover the level of prediction performance observed in models with demographic information. In particular, the teacher recommendation letters, which presumably contain gender pronouns, stand out by increasing the proportion of female applicants in the top pool significantly beyond the level in the model with demographic features $(\chi^2(df=1)=6.86, p=0.008)$.

While the addition of textual features recovers some of the loss in gender diversity from the omission of demographic features, we do not observe the same for racial diversity: the percentage of URM applicants drops substantially and none of the textual features substantially improve it. All textual materials including the writing supplement, which grants applicants an opportunity to write about their identity, slightly but not significantly raises the percentage of URM applicants in the top pool $(\chi^2(df=1)=2.47, p=0.12)$. The addition of teacher recommendation letters, which improved gender diversity, reduces the percentage of URM applicants but not significantly $(\chi^2(df=1)=0.12,  p=0.73)$. 

\begin{table}[!]
\centering
\caption{Percentage of female, URM, and admitted applicants in the top pool based on models trained on different combinations of textual and demographic features. Each model produces a ranking of applicants and the percentages indicate the composition of 10\% of applicants ranked highest by the model.}\label{tab2}
\begin{tabular}{|l|ccc|ccc|}
\hline
\multicolumn{1}{|c|}{ }  & \multicolumn{6}{c|}{\bfseries{Demographic Features}}\\
\multicolumn{1}{|l|}{\textbf{Textual Features}}  & \multicolumn{3}{c|}{Excluded} & \multicolumn{3}{c|}{Included}\\
\hline
& \multicolumn{1}{l|}{Female} & \multicolumn{1}{l|}{URM} & Admitted & \multicolumn{1}{l|}{Female} & \multicolumn{1}{l|}{URM} & Admitted \\ 
\hline
No text & \multicolumn{1}{c|}{45.8} & \multicolumn{1}{c|}{16.1} & 
19.1 & \multicolumn{1}{c|}{46.8} & \multicolumn{1}{c|}{64.2} & 
29.1 \\  
\hline
All text & \multicolumn{1}{c|}{48.1} & \multicolumn{1}{c|}{21.4} & 18.7 & \multicolumn{1}{c|}{46.4} & \multicolumn{1}{c|}{64.2} & 29.4 \\ 
\hline
Personal essay & \multicolumn{1}{c|}{41.8} & \multicolumn{1}{c|}{16.7} & 
18.4  & \multicolumn{1}{c|}{46.8} & \multicolumn{1}{c|}{64.2} & 
28.4 \\
\hline
Writing supplements & \multicolumn{1}{c|}{41.5} & \multicolumn{1}{c|}{20.7} & 
21.1 & \multicolumn{1}{c|}{44.8} & \multicolumn{1}{c|}{64.9} & 
27.6 \\
\hline
Teacher rec. letter & \multicolumn{1}{c|}{57.5} & \multicolumn{1}{c|}{14.7} & 
16.1 & \multicolumn{1}{c|}{46.5} & \multicolumn{1}{c|}{61.5} & 
28.9 \\
\hline
\end{tabular}
\end{table}

\begin{figure}
\centering
\includegraphics[scale=0.45]{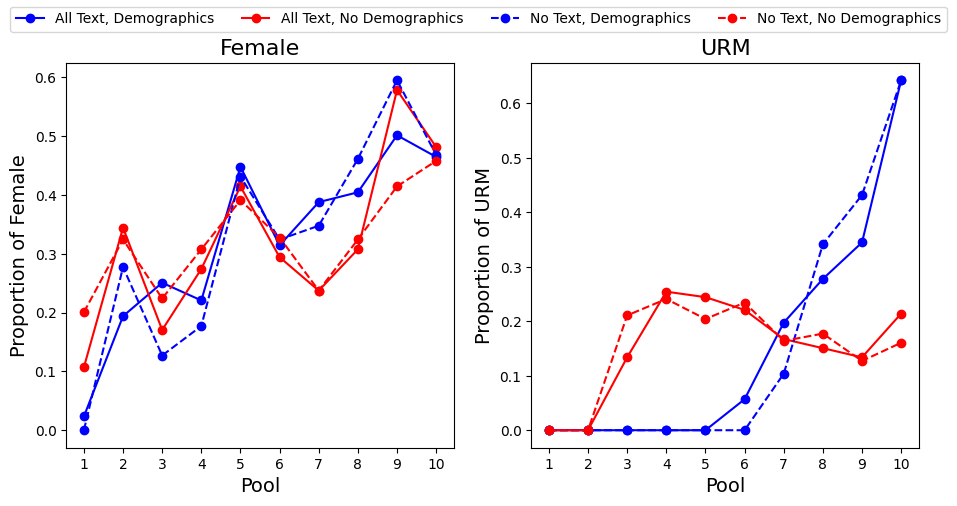}
\caption{Proportion of female (left) and URM (right) applicants in each of the 10 pools constructed from the prediction models with different combinations of demographic and textual features.}
\label{Fig:1}
\end{figure}

To examine these results beyond the top pool, Figure~\ref{Fig:1} illustrates the amount of diversity in terms of the proportion of female and URM applicants within each pool, comparing the models with and without demographic features to the models with all textual and without any textual features. We observe that across most applicant pools diversity drops when demographic features are omitted and the addition of textual features provides almost no improvement to diversity. We further observe that the percentage of URM applicants increases in pools with more highly ranked applicants for models that include demographic features, but it stays relatively constant for models without them.

\section{Discussion}

This study evaluated the consequences of changing the features used to develop a learned model for university admission designed to support a holistic review process. We considered a comprehensive range of factors within applications, including student identities, grades, honors, extracurricular activities, essays, writing supplements, and teacher recommendation letters. By examining prediction performance alongside gender and racial diversity, this research offers new insights for algorithmic fairness in the admission context.

The findings suggest that textual information in applications has the potential to enhance model prediction performance, especially when demographic data is unavailable. However, it is insufficient to make up for the loss of demographic features altogether, both in terms of prediction performance, and gender and racial diversity. The recommendation letter, presumably due to the explicit use of gender pronouns, was effective at recovering the loss in gender diversity when omitting demographic features. However, none of the texts were effective at recovering the substantial drop in the proportion of URM applicants. Unlike prior work that has found little effect of excluding protected attributes from student dropout prediction models on performance and fairness~\cite{yu2021should}, our findings underscore the significance of demographic attributes for accurately predicting admission decisions and enabling human-AI value alignment in terms of matching historical trends in admission decisions. 

The improvement in model performance from adding all textual features to the model with demographics was surprisingly minimal (AUC increased from .884 to .892) and statistically not significant ($p=0.303$). Essays and recommendation letters are thought to be rich sources of information that go beyond the limits of other fields in the application form. Yet the small performance gains we observe may indicate that there is little additional information contained in the textual information that is not already conveyed through grades, standardized test scores, extracurricular activities, honors, and awards. This finding echoes the conclusion of Alvero and colleagues who found that the content of admission essays correlates highly with standardized test scores and household income~\cite{alvero2021essay}. However, one has to keep in mind that the relatively simple textual representations used in the ML model are bound to miss some of the nuances that human reviewers can detect. 

This study has a number of limitations that need to be considered in the interpretation of its findings. First, the work was conducted at a single, highly selective institution; some of our findings may therefore not translate to other contexts that apply different admission criteria. Second, our results show high-level trends in prediction performance and diversity, but a more granular investigation of topics or words that contribute to the decisions for different groups is needed to understand why different features improve or deteriorate model performance and diversity. A deeper analysis of textual information can provide insights into how linguistic factors align with applicants' socio-demographic characteristics and how they express their identities. Finally, our representation of the textual features is limited to two relatively simple methods: TF-IDF and LDA topic modeling. LLMs such as GPT-4 or transformer-based models could improve prediction performance by providing a more nuanced and holistic representation of an applicant's identity, prior experiences, and writing style.

To achieve more informed and inclusive decision-making practices, this study underscores the importance of considering a comprehensive range of features, including applicant demographics, to effectively support administrators in college admission. Our findings caution against the notion that a loss in access to demographic features, especially race and ethnicity, can be addressed by using NLP and ML methods. To maintain similar levels of demographic diversity in the admitted pool without access to demographics, college admission professionals will likely need to rely on seasonal application readers to recover this information manually.
\raggedbottom
%
%
%
 \bibliographystyle{splncs04}
 \bibliography{nlp_admission}
\end{document}